\documentclass[11pt]{article}
\usepackage{coling2020}
\usepackage{times}
\usepackage{url}
\usepackage{latexsym}

\colingfinalcopy 

\usepackage{microtype}

\usepackage{hyperref}
\usepackage{siunitx} 

\usepackage{qtree}
\usepackage{forest}

\usepackage{float}
\usepackage{graphicx} 
\usepackage{amsthm, amsmath}
\usepackage{amssymb}
\usepackage{latexsym}

\usepackage{algpseudocode}
\usepackage{algorithm}

\usepackage{booktabs}
\usepackage{multirow}

\usepackage{caption}
\usepackage{subcaption}

\usepackage{bussproofs} 
\usepackage{stackengine}

\newcommand{\sortof}[1]{`#1'}
\newcommand{\of}[1]{\left(#1\right)}

\newlength\mylen

\makeatletter
\def\th@definition{%
  \thm@notefont{}
  \normalfont 
}
\makeatother

\theoremstyle{definition} 

\theoremstyle{definition}

\usepackage{amssymb}

\usepackage{pifont}

\usepackage[pdf,singlefile]{graphviz}

\usepackage{setspace}

\usepackage{color}
\usepackage{bm}
\definecolor{orange}{rgb}{1,0.5,0}
\definecolor{mdgreen}{rgb}{0.05,0.6,0.05}
\definecolor{mdblue}{rgb}{0,0,0.7}
\definecolor{dkblue}{rgb}{0,0,0.5}
\definecolor{dkgray}{rgb}{0.3,0.3,0.3}
\definecolor{slate}{rgb}{0.25,0.25,0.4}
\definecolor{gray}{rgb}{0.5,0.5,0.5}
\definecolor{ltgray}{rgb}{0.7,0.7,0.7}
\definecolor{purple}{rgb}{0.7,0,1.0}
\definecolor{lavender}{rgb}{0.65,0.55,1.0}
\definecolor{brown}{rgb}{0.6,0.2,0.2}

\newcommand{\code}[1]{}

\usepackage{bm}
\usepackage{xspace}

\newcommand{\app}[1]{\text{\textsc{App}}\ensuremath{_{#1}}}  
\newcommand{\modify}[1]{\text{\textsc{mod}}\ensuremath{_{#1}}\xspace}

\newcommand{\src}[1]{\text{\textsc{#1}}\xspace}                                  
\newcommand{\obj}[1][]{\text{\src{o}}\ensuremath{_{#1}}}               

\newcommand{\G}[1]{\ensuremath{G_{\text{#1}}}} 

\newcommand{\el}[1]{\text{`#1'}}

 \newcommand{\word}[1]{``#1''}
  \newcommand{\sentence}[1]{``#1''}

\newcommand{\ci}[1]{\scalebox{0.8}{\ensuremath{\pm #1}}}

\usepackage{tikz}
\usepackage{tikz-qtree}

\usetikzlibrary{graphs}
\usetikzlibrary{shapes,arrows}
\usetikzlibrary{positioning}
\usetikzlibrary{quotes}

\tikzset{snode/.style={
    ellipse,
    minimum size=6mm,
    very thick,
    draw=black,
    font=\rmfamily}}

\tikzset{ssrc/.style={
    ellipse,
    minimum size=6mm,
    very thick,
    fill=black!10,
    draw=black,
    text=black,
    font=\rmfamily}}

\tikzset{sanno/.style={node distance=-1mm,font=\rmfamily\small}}

\tikzset{sedgel/.style={
    color=black,
    sloped, above,
    font=\rmfamily\small}}

\tikzset{sedge/.style={
    thick,
    >=stealth'
}}

\usepackage{array}

\usepackage{tikz-dependency} 
\depstyle{amdep}{label style={text=black, fill=black!10}, edge style={thick}, edge slant=1pt}
\depstyle{sdpgraph}{text only label, label style={font=\small\slshape, above}, edge style={>=latex,solid,gray}, edge slant=5pt}

\newcommand{\pattern}[1]{\textsc{#1}}
\newcommand{\pzero}{\pattern{empty}} 
\newcommand{\pone}{\pattern{app2}}
\newcommand{\ptwo}{\pattern{conn}}
\newcommand{\pthree}{\pattern{capp2}}
\newcommand{\pfour}{\pattern{cmod}}
\newcommand{\pfive}{\pattern{app1}}
\newcommand{\psix}{\pattern{mod}}
\newcommand{\pseven}{\pattern{basic}}
\newcommand{\peight}{\pattern{other}}
\newcommand{\patterntriple}[3]{[#1 / #2 / #3]}

\newcommand{\pos}[1]{\textit{#1}}

\newcommand{\lnt}{L19}
\newcommand{\lntree}{\lnt\ AM tree}

\title{Normalizing Compositional Structures Across Graphbanks}

\author{Lucia Donatelli, Jonas Groschwitz, Alexander Koller,\AND Matthias Lindemann, Pia Wei\ss enhorn \\
  Department of Language Science and Technology \\
  Saarland University, Germany \\
  {\tt \{donatelli,jonasg,koller,mlinde,piaw\}@coli.uni-saarland.de}}

\date{}

\begin{document}
\maketitle
\begin{abstract}
  The emergence of a variety of graph-based meaning representations (MRs) has sparked an important conversation about how to adequately represent semantic structure. These MRs exhibit structural differences that reflect different theoretical and design considerations, presenting challenges to uniform linguistic analysis and cross-framework semantic parsing. Here, we ask the question of which design differences between MRs are meaningful and semantically-rooted, and which are superficial. We present a methodology for normalizing discrepancies between MRs at the compositional level \cite{lindemann2019compositional}, finding that we can normalize the majority of divergent phenomena using linguistically-grounded rules. Our work significantly increases the match in compositional structure between MRs and improves multi-task learning (MTL) in a low-resource setting, demonstrating the usefulness of careful MR design analysis and comparison.    
\end{abstract}

\section{Introduction}\label{sec:intro}

Graph-based representations of sentence meaning offer an expressive
and flexible means of modeling natural language semantics. In recent
years, a number of different \emph{graphbanks} have annotated large
corpora with graph-based semantic representations of various types, for example Elementary Dependency Structures (EDS) \cite{oepen-lonning-2006-eds}; DELPH-IN MRS Bi-Lexical Dependencies (DM) \cite{ivanova-etal-2012-dm}; Abstract Meaning Representation (AMR) \cite{banarescu-etal-2013-amr}; and Universal Conceptual Cognitive Annotation (UCCA) \cite{abend-rappoport-2013-ucca}. Because of differences in the design
principles underlying the different graphbanks, these graphs differ
greatly and often in fundamental strategies such as how lexical items are anchored in graph nodes \cite{oepen-etal-2019-mrp}. 
As an example, the graphs shown in Fig.~\ref{fig:catdoghouseSDP} correspond to the three graphbanks of the SemEval 2015 Shared Task on Semantic Dependency Parsing, which we take as our focus in this paper. The graphs visibly
differ with respect to edge labels, edge directions, the treatment of
a periphrastic verb construction, and coordination. 

Understanding the nature of these differences is important both from a
linguistic and from a parsing perspective. With respect to
linguistics, one can ask which design differences between graphbanks
are meaningful (e.g. understanding coordinate structures as headed by a conjunct or the conjunction); which ones are artifacts that can be equalized by
simple transformations (e.g.\ flipping the direction of edges); and
which ones are worth spending time on when discussing the design of
future corpora. With respect to parsing, the differences between
graphbanks make it hard to develop parsers that are accurate across
many graphbanks \cite{oepen-etal-2019-mrp}. A more unified approach to certain structures
across graphbanks has the potential to facilitate multi-task learning (MTL)
to train parsers on smaller graphbanks by exploiting the training set
of a larger graphbank, in addition to being linguistically parsimonious.

In this paper, we show how annotations from different graphbanks can
be normalized at the level of their compositional structures. We build
upon the work of \newcite{lindemann2019compositional}, who used 
\emph{AM dependency trees} to represent the compositional structure of
graph-based meaning representations (MRs); they developed graphbank-specific heuristics
for automatically constructing AM dependency trees for the graphs in a
number of different graphbanks built upon the \textit{AM algebra} \cite{groschwitz2018amr} (Section~\ref{sec:background}). We develop a new methodology for identifying and
quantifying mismatches between the compositional structures assigned
to the same sentence by different graphbanks;  
we then present our extended \textit{AM+ algebra}, which is able to systematically reshape these compositional structures in order to make
them more uniform across graphbanks (Section \ref{sec:updating}).
Finally, we walk through how our methodology can be applied to normalize specific linguistic phenomena that differ in representation across graphbanks (Section \ref{subsec:unifyingkeyphenom}). Our work is focused on the three graphbanks
of the SemEval 2015 Shared Task on Semantic Dependency Parsing
\cite{oepen-etal-2015-semeval} -- DM, PAS, and PSD -- as these
are annotated over the same base text corpus. Our methods can in principle be applied to more complex graphbanks, as well.

Using our methods, we increase the match between the AM dependency
trees (compositional structure) of the different graphbanks to 76.3 (DM-PAS), 78.8 (DM-PSD), and 82.0
(PAS-PSD) directed unlabeled F-score, compared to 63.5, 55.7, and 57.0 for Lindemann
et al.'s AM dependency trees (Section~\ref{sec:evaluation}). This is a drastic improvement over the
unlabeled F-scores of 64.2, 26.1, and 29.6 for the original graphs
\cite{oepen-etal-2015-semeval}. We additionally demonstrate that when
training a graph parser on a tiny graphbank combined with larger
graphbanks using multi-task learning (MTL), our methods improve parsing
accuracy by up to 2.6 points F-score over MTL without normalized AM dependency trees. Finally, we assess the hierarchy of design differences with respect to their theoretical and quantitative impacts and discuss phenomena which cannot be made uniform
using the techniques developed here (Section~\ref{sec:discussion}).

\section{Background}\label{sec:background}

We first describe the graphbanks we use for our experiment in this paper (Section \ref{subsec:graphbanks}). We then motivate our use of the compositional AM algebra and parser that we use for our implementation (Section \ref{subsec:graphbankparsing}).

\subsection{The Graphbanks}\label{subsec:graphbanks}


\begin{figure}
  \centering
  \begin{subfigure}{\textwidth}
    \centering
    \includegraphics{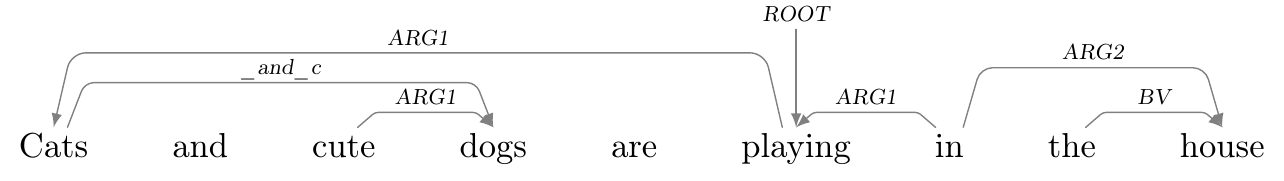}
    \caption{DM}\label{fig:catdoghouseSDP:DM}
  \end{subfigure}

  \begin{subfigure}{\textwidth}
    \centering
    \includegraphics{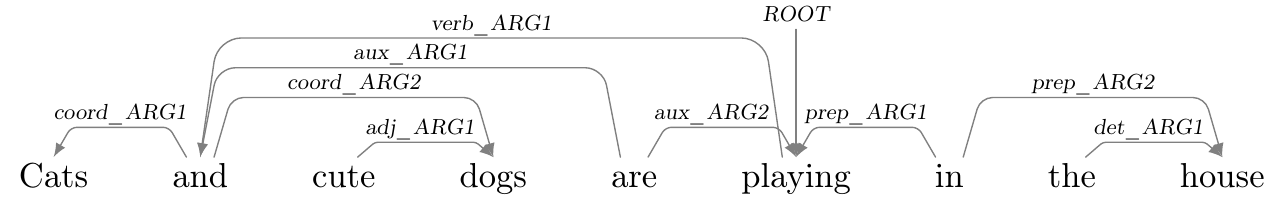}
    \caption{PAS}\label{fig:catdoghouseSDP:PAS}
  \end{subfigure}

  \begin{subfigure}{\textwidth}
    \centering
    \includegraphics{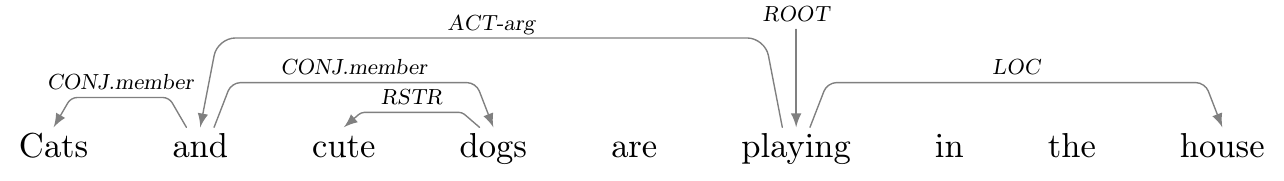}
    \caption{PSD}\label{fig:catdoghouseSDP:PSD}
  \end{subfigure}
  \caption{Examples for DM, PAS and PSD graphs. The PSD graph includes Lindemann et al.'s (2019) reversible preprocessing of coordination.
  }\label{fig:catdoghouseSDP}
\end{figure}

We focus our efforts on the three graphbanks of the SemEval 2015 Shared Task on Semantic Dependency Parsing (SDP): (i) DELPH-IN MRS-Derived Semantic Dependencies (DM), (ii) Enju Predicate–Argument Structures (PAS), and (iii) Prague Semantic Dependencies (PSD) \cite{oepen-etal-2015-semeval}. All graphbanks have parallel semantic annotations over the same common English texts: the Wall Street Journal (WSJ) and Brown segments of the Penn Treebank (PTB; \newcite{marcus-etal-1993-building}). These graphbanks are all bilexical, node-ordered, labeled, directed graphs and represent core semantic information such as predicate-argument relations. Nothing in our paper hinges on the choice of these particular graphbanks; the methods presented here can be applied to other graphbanks, though we assume that the graphs are over the same base texts.

The SDP graphbanks make different choices about which linguistic
information to represent and how to represent it
\cite{oepen2016towards} (Table \ref{tab:similarites}). Graphs in the style
of the three graphbanks for the sentence \sentence{Cats and cute dogs are
playing in the house} are shown in Fig.~\ref{fig:catdoghouseSDP}.
While in all three graphs each token of the
sentence corresponds to at most one node, there may also be tokens
which are not part of the graph. These \sortof{ignored} tokens
differ across graphbanks; for instance, in DM
coordinators (\word{and}) and temporal auxiliaries (\word{are}) are not nodes in the graph, while PSD ignores
temporal auxiliaries, prepositions, and determiners. The edges are also different;
for instance, edges that point from adjectives to nouns in DM and PAS
go the other way in PSD.

\begin{table}[]
    \centering

\begin{tabular}{ccccc}
\toprule
& \multicolumn{2}{c}{Directed F} & \multicolumn{2}{c}{Undirected F} \\
& PAS & PSD & PAS & PSD \\
\midrule
DM & 64.2 & 26.1 & 67.2 & 56.8 \\
PAS & & 29.6 & & 54.9 \\
\bottomrule
\end{tabular}
\caption{Unlabeled F-scores across formalisms including punctuation according to \protect\newcite{oepen-etal-2015-semeval}.}
\label{tab:similarites}
\end{table}

The fundamental question we discuss in this paper is the extent to
which these variations are superficial or, alternatively, represent
irreconcilable design differences. To this end, we make the standard
assumption from theoretical semantics that MRs are
derived \emph{compositionally} from the sentence, i.e.\ by
systematically computing MRs for each part of the sentence from
its immediate subparts. We say that a difference between two graphs is
\sortof{superficial} if it can be derived using the same compositional
structure. To give an example, Fig.~\ref{fig:catdoghouseAmdep}(\subref{fig:catdoghouseAmdepDM},~\subref{fig:catdoghouseAmdepPSD}) show the compositional structure which
\newcite{lindemann2019compositional} assign to the DM and PSD graphs
in Fig.~\ref{fig:catdoghouseSDP}. We explain the technical
details in Section~\ref{subsec:graphbankparsing}; for now, observe
that certain differences between DM and PSD have been normalized
(e.g.\ both dependency trees analyze the noun as the head and the
adjective as the modifier), while others are still present (e.g.\
ignored tokens). In this paper, we present a method for deriving
compositional structures for SDP sentences that are more
consistent across graphbanks and still linguistically solid. This
allows us to distinguish superficial design differences more cleanly
from deeper ones, and is useful for parsing.


\subsection{Compositional Semantic Parsing with the AM Algebra}\label{subsec:graphbankparsing}

To normalize discrepancies between the three formalisms, we use the \textit{Apply-Modify (AM) Algebra} initially presented in \newcite{groschwitz2018amr} for compositional AMR parsing. Utilizing the AM algebra allows us to modify graph structure at the compositional level, represented as \textit{AM dependency trees} (explained below). \newcite{lindemann2019compositional} (hereafter referred to as \lnt) implemented this methodology and extended the AM algebra to four additional graph-based MRs (DM, PAS, PSD, and EDS); \newcite{donatelli2019saarland} extended this work to UCCA. \lnt\ achieved state of the art results for all formalisms in question with pretrained BERT embeddings and manual heuristics tailored to each graphbank. We adapt these heuristics to find uniform structures that generalize across graphbanks.  

\begin{figure}
	\centering
	\begin{subfigure}[b]{0.45\linewidth} 
		\centering
		\includegraphics[scale=1.1]{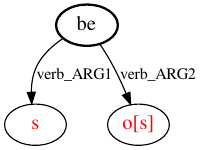}\hspace{-5pt}
		\includegraphics[scale=0.8]{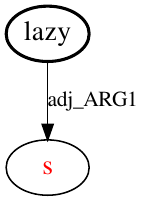}\hspace{-10pt}
		\includegraphics[scale=0.4]{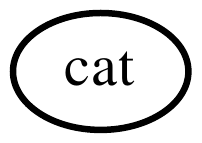}
		\includegraphics[scale=0.8]{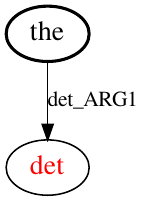}\hspace{-10pt}
		\includegraphics[scale=0.8]{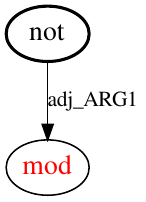}
		\caption{Constants \G{be}, \G{lazy}, \G{cat}, \G{the}, and \G{not}}\label{fig:am:consts}
	\end{subfigure}
	\begin{subfigure}[b]{.25\linewidth}
		\centering
		\includegraphics[scale=0.35]{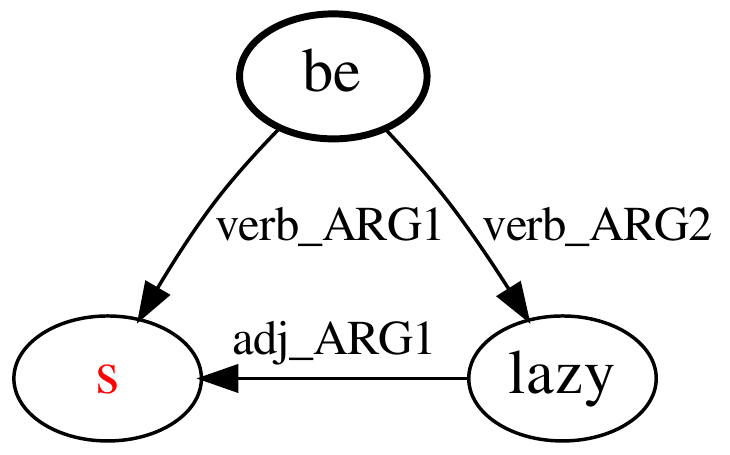}
		\caption{$\app{\src{O}}\of{\G{be},\G{lazy}}$}\label{fig:am:beLazy}
	\end{subfigure}
	\begin{subfigure}[b]{.25\linewidth}
		\centering
		\includegraphics[scale=0.35]{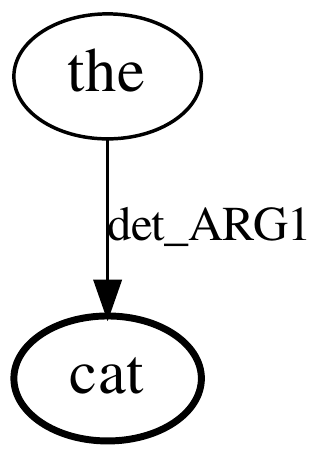}
		\caption{$\modify{\src{det}}\of{\G{cat},\G{the}}$}\label{fig:am:theCat}
	\end{subfigure}
	
	\hspace{-15mm}
	\begin{subfigure}[b]{0.25\linewidth} 
		\centering
		\includegraphics[scale=0.5]{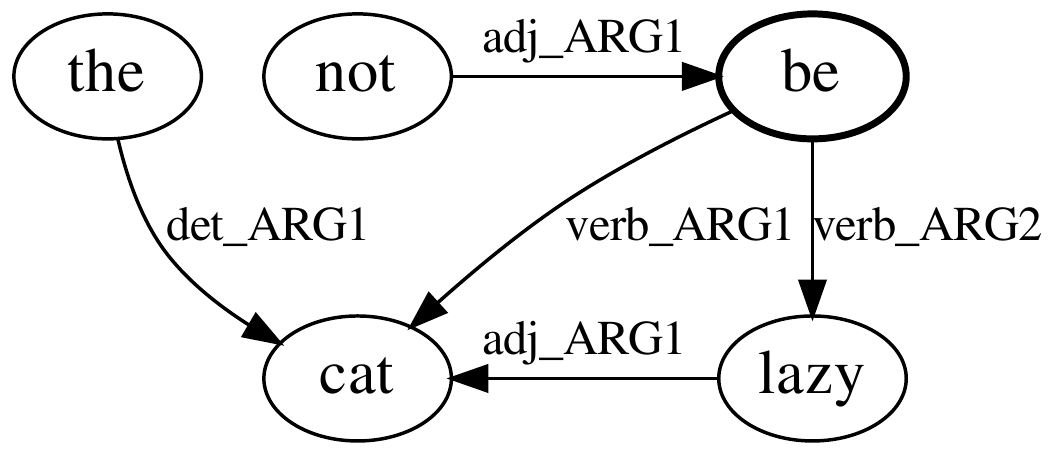}
		\caption{PAS graph (cf.\ Fig.~\ref{fig:copulaPASgraph})}\label{fig:am:graph}
	\end{subfigure}
	\begin{subfigure}[b]{.35\linewidth}
		\begin{center}
			\resizebox{.7\linewidth}{!}{
			\begin{forest}
				for tree={align=center, l=\baselineskip, l sep=5pt} 
				[{\modify{\src{mod}}}, for tree={fill=green!30}
					[{\app{\src{s}}}, for tree={fill=green!30}
						[{\app{\obj}}
							[{\G{be}},tier=word]
							[{\G{lazy}}, for tree={fill=red!30},tier=word]
						]
						[{\modify{\src{det}}}, for tree={fill=brown!10!yellow!60}
							[{\G{cat}},tier=word]
							[{\G{the}}, for tree={fill=blue!30},tier=word]
						]
					]
					[{\G{not}}, for tree={fill=orange!30},tier=word]
				]
			\end{forest}}
		\end{center}\vspace{-.5em}
		\caption{AM term}\label{fig:am:term}
	\end{subfigure}\hspace{-9mm}
	\begin{subfigure}[b]{0.37\linewidth}
			\centering
			\resizebox{1.2\linewidth}{!}{
			\begin{dependency}[amdep, label style={font=\Large}]
				\begin{deptext}[column sep=.3cm, row sep=0.2ex]
					The \& cat \& is \& not \&[-2mm] lazy \\
					\fcolorbox{black}{blue!30}{\G{the}} \& \fcolorbox{black}{brown!10!yellow!60}{\G{cat}} \& \fcolorbox{black}{green!30}{\G{be}} \& \fcolorbox{black}{orange!30}{\G{not}} \& \fcolorbox{black}{red!30}{\G{lazy}} \\ 
					\& \& \& \& \\
				\end{deptext}
				\deproot[edge unit distance=2ex]{3}{root}
				\depedge[edge unit distance=2ex, label style={fill=brown!10!yellow!60}]{2}{1}{\modify{\src{det}}} 
				\depedge[edge unit distance=2ex, label style={fill=green!30}]{3}{2}{\app{\src{s}}}      
				\depedge[edge unit distance=2.5ex, label style={fill=green!30}]{3}{5}{\app{\src{o}}}      
				\depedge[edge unit distance=2ex, label style={fill=green!30}]{3}{4}{\modify{\src{mod}}}    
			\end{dependency}
			} 
			\vspace{-2.5em}
		\caption{AM dependency tree}\label{fig:am:deptree}
	\end{subfigure}
	\caption{PAS for \sentence{the cat is not lazy} with its AM analysis according to \lnt . }\label{fig:am}
\end{figure}

The AM algebra builds
sentence-level graphs from graph fragments called \textit{annotated s-graphs}, or
\textit{as-graphs}. Fig.~\ref{fig:am:consts}
shows as-graphs from which the PAS graph in Fig.~\ref{fig:am:graph} for the
sentence \sentence{the cat is not lazy} can be built. Note that Fig.~\ref{fig:am:graph} is the same as the graph in Fig.~\ref{fig:copulaPASgraph} minus the ordering among the nodes.
As-graphs possess two kinds of markers: (i) an obligatory \emph{root source} on the \emph{root node} (indicated with bold outline); and (ii) optionally further \textit{sources} (marked in red, e.g.\ \src{s} and \src{o}) on other nodes, which allow us to combine as-graphs. Every node can bear at most one source, and each source can occur at most once in an as-graph. 

The AM algebra defines two operations for combining as-graphs. (i) \textit{Apply} for a source \src{X} (\app{\src{X}}) combines head and argument as-graphs by filling the head's
\src{X}-source with the root of the argument. For example, the result of $\app{\src{o}}\of{\G{be},\G{lazy}}$ is shown in Fig.~\ref{fig:am:beLazy}.\footnote{Note: \word{lazy} is represented in Fig.~\ref{fig:am:beLazy} with an \src{s}-source for ease of understanding here; \lnt\ analyze it as a \src{mod}-source. We use \src{s}-sources for adjectives in PAS in the presentation of all examples.}
Nodes in both graphs with the same source (here \src{s}) are merged. The type annotation $[\src{s}]$ at the \src{o}-source of \G{be} requests the argument to have an \src{s}-source (which \G{lazy} has). This way, type annotations lead to reentrancies in the graph (like the triangle structure here). 
(ii) \textit{Modify} allows an as-graph to modify a head. For example,
$\modify{\src{det}}\of{\G{cat},\G{the}}$ and its result is shown in
Fig.~\ref{fig:am:theCat}. Here, \G{cat} is the head and \G{the} the modifier; \G{the} attaches with its \src{det}-source at \G{cat} and loses its own root source. We combine the two partial results we have built so far using \app{\src{s}}. 
The final graph is the result of the \modify{\src{mod}} operation at the top of the term in
Fig.~\ref{fig:am:term}.



By tracking the \sortof{semantic heads} of
each subtree of an AM term as color coded in Fig.~\ref{fig:am:term}, \lnt\ encode
AM terms as AM dependency trees (Fig.~\ref{fig:am:deptree}):
whenever the AM term combines two graphs with some operation,
a dependency edge is added from one semantic head to the other.
The graph constants of the AM term become the \emph{lexical as-graphs} for
the tokens in the sentence (if a token has no graph constant associated with it,
we write `$\bot$'). Such an AM dependency
tree evaluates deterministically to a
graph. 

A central challenge with AM dependency parsing is that the AM dependency trees in the training corpus are latent. \lnt\ formulated graphbank-specific decomposition heuristics to address this.
In the context of normalizing the graphbanks, this challenge turns into an opportunity: we can normalize the graphbanks at their compositional level without having to directly change the graphs.
In fact, even the AM dependency trees of \lnt\ naturally normalize some differences of the MRs. Consider the example \sentence{is not} in the PAS and PSD graphs in Fig.~\ref{fig:copulagraphs}(\subref{fig:copulaPASgraph},~\subref{fig:copulaPSDgraph}) and their \lntree s in Fig.~\ref{fig:copulagraphs}(\subref{fig:copulaPASdecomp},~\subref{fig:copulaPSDdecomp}). In the graphs, the edges between \word{is} and \word{not} have a different label and different direction, whereas both AM trees have a \modify{\src{mod}} edge with the same direction. Furthermore, in the same graphs, PAS has a triangle \word{is}-\word{lazy}-\word{cat}, while PSD does not. In the \lntree s (Fig.~\ref{fig:copulagraphs}(\subref{fig:copulaPASdecomp},~\subref{fig:copulaPSDdecomp})), this difference is reduced to the lexical as-graphs.

\section{Quantifying Compositional Mismatches}\label{sec:updating}

\begin{figure}
	\centering
	\begin{subfigure}[b]{.3\textwidth}
		\centering
		\includegraphics[scale=0.9]{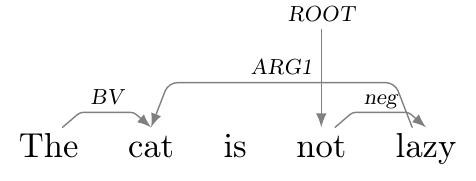}
		\caption{DM graph}\label{fig:copulaDMgraph}
	\end{subfigure}
	\begin{subfigure}[b]{.4\textwidth}
		\centering
		\includegraphics[scale=0.9]{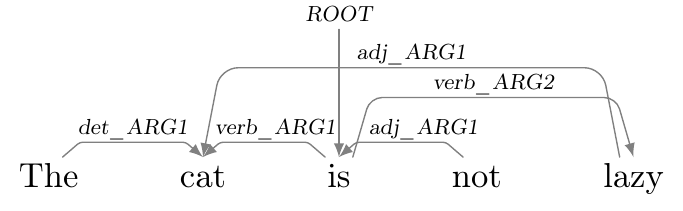}
		\caption{PAS graph}\label{fig:copulaPASgraph}
	\end{subfigure}
	\begin{subfigure}[b]{.25\textwidth}
		\centering
		\includegraphics[scale=0.9]{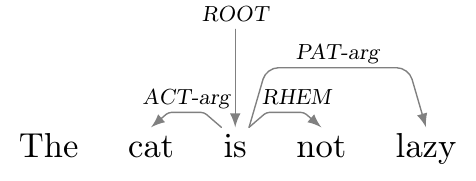}
		\caption{PSD graph}\label{fig:copulaPSDgraph}
	\end{subfigure}
	
	\begin{subfigure}[b]{.28\textwidth}
		\centering
		\includegraphics[scale=.7]{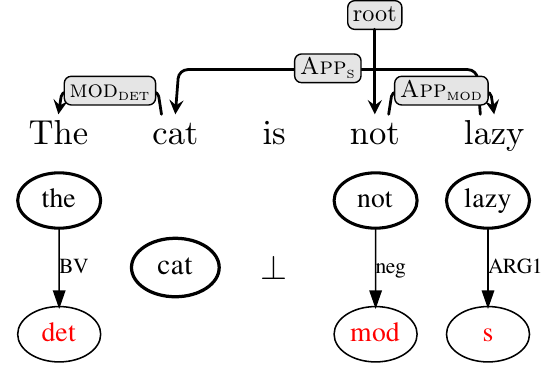}
		\caption{\lntree\ for DM}\label{fig:copulaDMdecomp}
	\end{subfigure}
	\begin{subfigure}[b]{.4\textwidth}
		\centering
		\includegraphics[scale=.7]{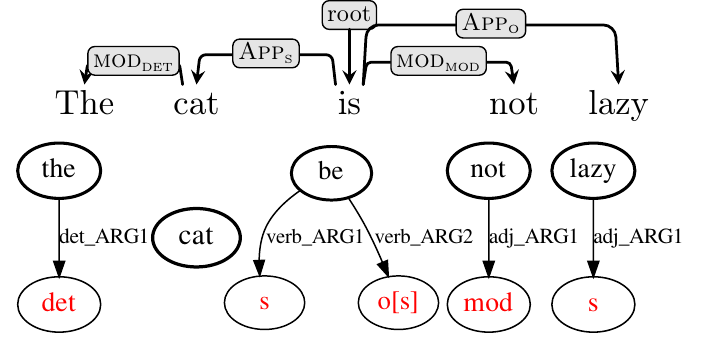}
		\caption{\lntree\  for PAS}\label{fig:copulaPASdecomp}
	\end{subfigure}
	\begin{subfigure}[b]{.28\textwidth}
		\centering
		\includegraphics[scale=.7]{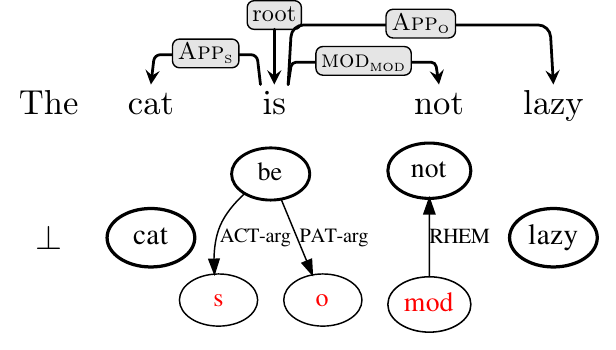}
		\caption{\lntree\  for PSD}\label{fig:copulaPSDdecomp}
	\end{subfigure}
	
	\begin{subfigure}[b]{.28\textwidth}
		\centering
		\includegraphics[scale=.7]{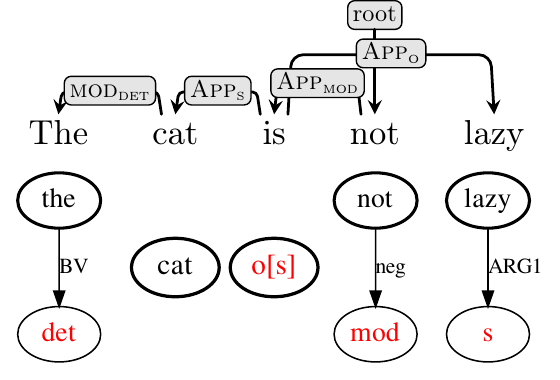}
		\caption{normalized tree for DM}\label{fig:copulaDMdecompNew}
	\end{subfigure}
	\begin{subfigure}[b]{.4\textwidth}
		\centering
		\includegraphics[scale=.7]{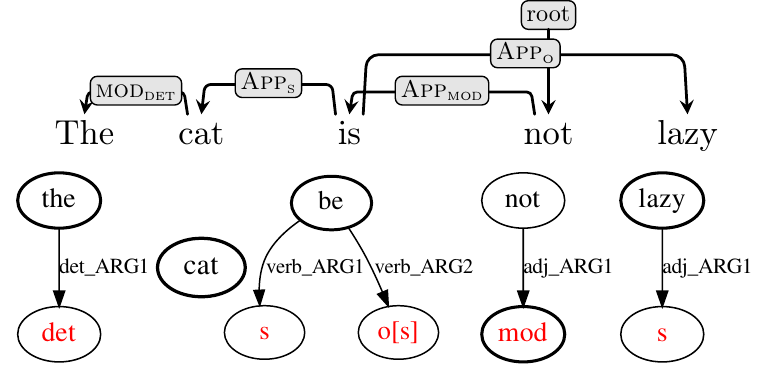}
		\caption{normalized tree for PAS}\label{fig:copulaPASdecompNew}
	\end{subfigure}
	\begin{subfigure}[b]{.28\textwidth}
		\centering
		\includegraphics[scale=.7]{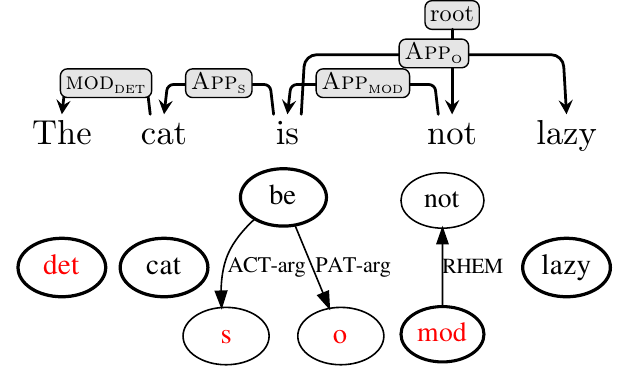}
		\caption{normalized tree for PSD}\label{fig:copulaPSDdecompNew}
	\end{subfigure}
	\caption{SDP graphs, \lntree s and normalized trees for the sentence \sentence{The cat is not lazy} 
	}\label{fig:copulagraphs}
\end{figure}

\begin{figure}
  \centering
  \begin{subfigure}[b]{.48\textwidth}
    \centering
    \includegraphics[scale=.7]{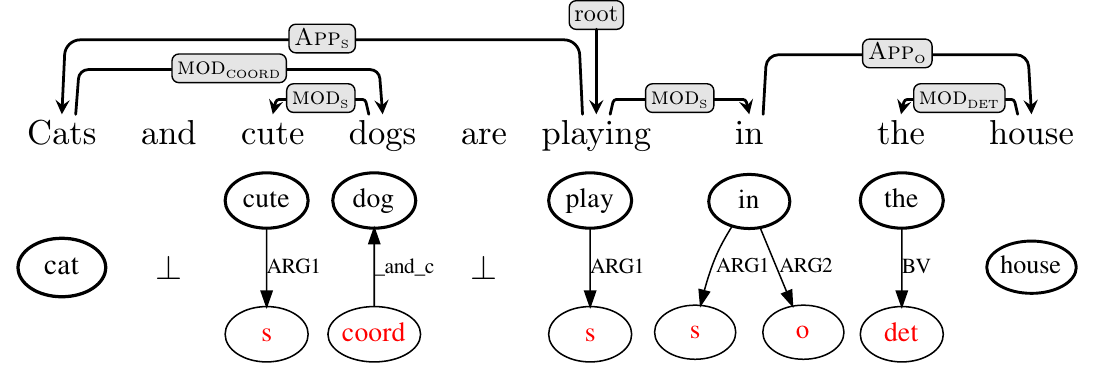}
    \caption{\lntree\  for DM}\label{fig:catdoghouseAmdepDM}
  \end{subfigure}
  \begin{subfigure}[b]{.48\textwidth}
    \centering
    \includegraphics[scale=.7]{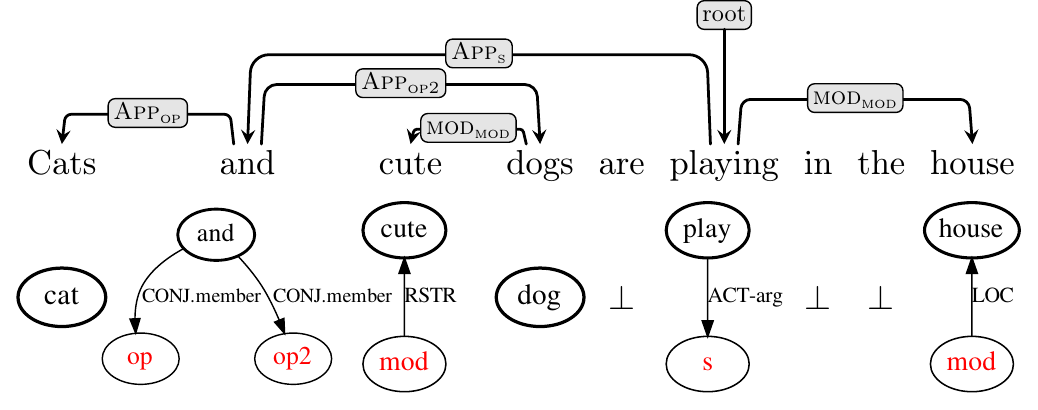}
    \caption{\lntree\  for PSD}\label{fig:catdoghouseAmdepPSD}
  \end{subfigure}

  \begin{subfigure}[b]{\textwidth}
    \centering
    \includegraphics[scale=.7]{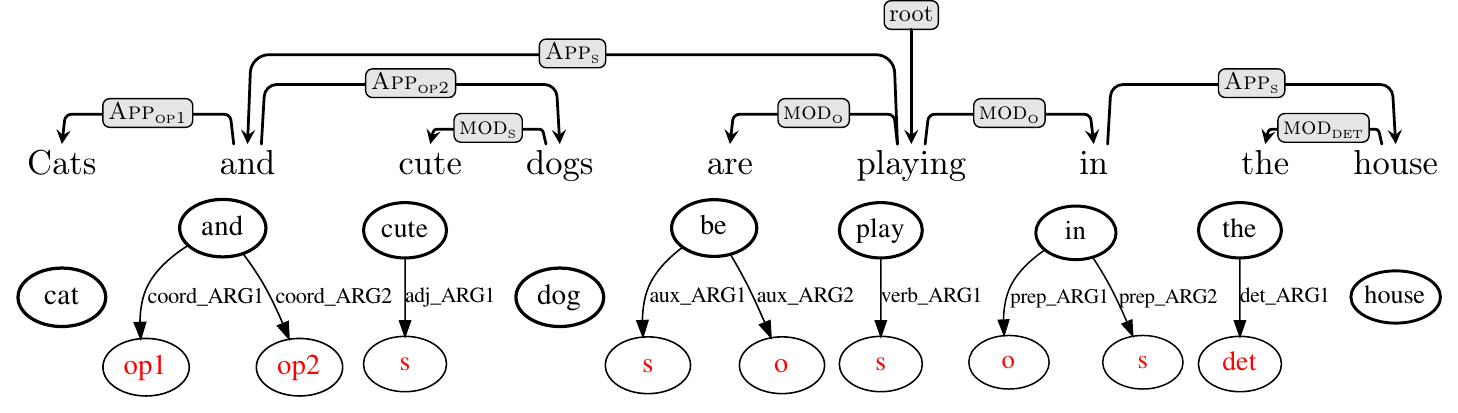}
    \caption{\lntree\  is equal to normalized tree for PAS}\label{fig:catdoghouseAmdepPAS}
  \end{subfigure}

  \begin{subfigure}[b]{.48\textwidth}
    \centering
    \includegraphics[scale=.7]{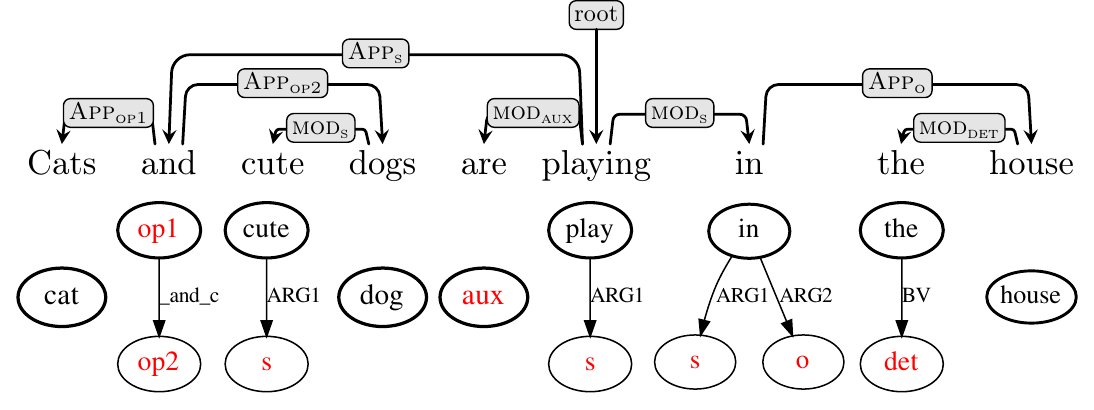}
    \caption{normalized tree for DM}\label{fig:catdoghouseAmdepDMnew}
  \end{subfigure}
  \begin{subfigure}[b]{.48\textwidth}
    \centering
    \includegraphics[scale=.7]{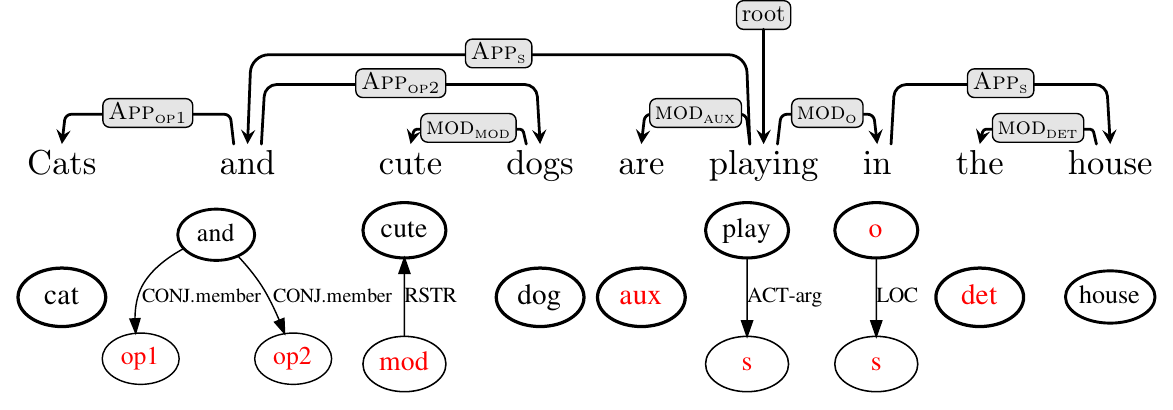}
    \caption{normalized tree for PSD}\label{fig:catdoghouseAmdepPSDnew}
  \end{subfigure}
  \caption{\lntree s and normalized trees for SDP graphs of \sentence{Cats and cute dogs are playing in the house} presented in Fig.~\ref{fig:catdoghouseSDP}.
  }\label{fig:catdoghouseAmdep}
\end{figure}

The AM dependency trees of \lnt\ (henceforth, \textit{\lntree s})
constitute a first step towards normalizing MRs
across graphbanks by making compositional structures explicit. This has a positive result: the AM
dependency trees in Fig.~\ref{fig:catdoghouseAmdep}(\subref{fig:catdoghouseAmdepDM}-\subref{fig:catdoghouseAmdepPAS}) are more
consistent with each other than the original graphs in
Fig.~\ref{fig:copulagraphs}. Yet, there remain differences between the \lntree s. The question we ask here is, can we find a single compositional structure which will correspond to all three MR graphs by changing only the lexical as-graphs for the individual tokens and not the \lntree s themselves? This would suggest that
the designs and annotators of the three graphbanks tacitly agree on
compositional structure; surface differences between the graphs can then be explained by low-level representations of
individual words.

In this section, we present our methodology to apply transformations to the \lntree s that do not change the graphs they evaluate to. We show how to detect and quantify mismatches
between the \lntree s for the same sentence in terms of \emph{pattern
  signatures} (Section~\ref{subsec:localpatterns}). As we will then
see, the AM algebra has limitations which prevent us from fully
normalizing the compositional structures. We will therefore slightly
extend it to the \emph{AM+ algebra}
(Section~\ref{subsec:amplusalgebra}), which we will use in
Section~\ref{subsec:unifyingkeyphenom} to automatically normalize AM
dependency trees, making them more consistent with each other.


\subsection{Detecting Local Patterns}\label{subsec:localpatterns}


\begin{figure}
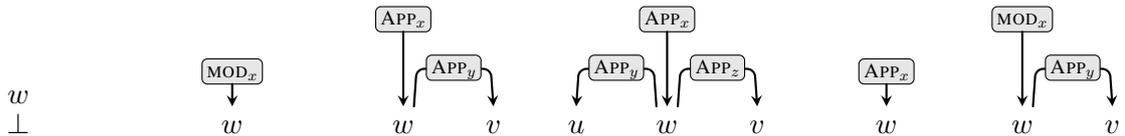

	\centering
	\begin{subfigure}[b]{.15\textwidth}
		\centering
		\begin{dependency}[amdep]
			\begin{deptext} $w$ \\ $\bot$ \\ \end{deptext}
		\end{dependency}
		\caption{Pattern \pzero}\label{fig:patterns:zero}
	\end{subfigure}
	\begin{subfigure}[b]{.19\textwidth}
		\centering
		\begin{dependency}[amdep]
			\begin{deptext} $w$ \\ \end{deptext}
			\deproot[edge unit distance=1ex]{1}{\modify{x}}
		\end{dependency} 
		\caption{Pattern \psix/\pfour}\label{fig:patterns:sixfour}
	\end{subfigure}
	\begin{subfigure}[b]{.15\textwidth}
		\centering
		\begin{dependency}[amdep]
			\begin{deptext}[column sep=2em] $w$ \& $v$\\ \end{deptext}
			\deproot[edge unit distance=2ex]{1}{\app{x}}
			\depedge{1}{2}{\app{y}}
		\end{dependency} 
		\caption{Pattern \pfive}\label{fig:patterns:five}
	\end{subfigure}
	\begin{subfigure}[b]{.2\textwidth}
		\centering
		\begin{dependency}[amdep]
			\begin{deptext}[column sep=2em] $u$ \& $w$ \& $v$\\ \end{deptext}
			\deproot[edge unit distance=2ex]{2}{\app{x}}
			\depedge{2}{1}{\app{y}}
			\depedge{2}{3}{\app{z}}
		\end{dependency} 
		\caption{Pattern \pone/\pthree}\label{fig:patterns:onethree}
	\end{subfigure}
	\begin{subfigure}[b]{.15\textwidth}
		\centering
		\begin{dependency}[amdep]
			\begin{deptext} $w$ \\ \end{deptext}
			\deproot[edge unit distance=1ex]{1}{\app{x}}
		\end{dependency} 
		\caption{Pattern \pseven}\label{fig:patterns:seven}
	\end{subfigure}
	\begin{subfigure}[b]{.13\textwidth}
		\centering
		\begin{dependency}[amdep]
			\begin{deptext}[column sep=2em] $w$ \& $v$\\ \end{deptext}
			\deproot[edge unit distance=2ex]{1}{\modify{x}}
			\depedge{1}{2}{\app{y}}
		\end{dependency} 
		\caption{Pattern \ptwo}\label{fig:patterns:two}
	\end{subfigure}
	\caption{The set of local patterns we distinguish based on their structures in L19 AM trees. 
	}\label{fig:patterns}
\end{figure}


In order to normalize the compositional structures of the graphbanks automatically, we must be able to detect and quantify discrepancies between their \lntree s automatically. We do this by identifying \sortof{local} patterns in the \lntree s trees that correspond to individual tokens (Fig.~\ref{fig:patterns}). We then compare these local patterns across the \lntree s for the three graphbanks to uncover a \textit{pattern signature}, with which we can identify and normalize differences across the graphbanks. 

To illustrate how this works, take the case of the determiner \word{the} in Fig.~\ref{fig:copulagraphs}. In the graphs themselves (\subref{fig:copulaDMgraph}--\subref{fig:copulaPSDgraph}), DM (\subref{fig:copulaDMgraph}) and PAS (\subref{fig:copulaPASgraph}) mark the determiner with different relations while PSD (\subref{fig:copulaPSDgraph}) ignores it. This is reflected in \lntree s (\subref{fig:copulaDMdecomp}--\subref{fig:copulaPSDdecomp}): the DM (\subref{fig:copulaDMdecomp}) and PAS (\subref{fig:copulaPASdecomp}) trees combine the determiner with a \modify{} relation, while the PSD tree (\subref{fig:copulaPSDdecomp}) leaves it absent ($\bot$). We name these patterns \psix~for DM and PAS, and \pzero~for PSD, and create a pattern signature with the triple \patterntriple{DM}{PAS}{PSD}:  \patterntriple{\psix}{\psix}{\pzero}. We use this order for all pattern signatures. Table~\ref{tab:patternSignatures} shows that the determiner pattern signature is the most common across the MRs, alongside the other most frequent pattern signatures we find and fix.


The set of patterns shown in Fig.~\ref{fig:patterns} are the basis for the pattern signatures we utilize to capture a range of linguistic phenomena.\footnote{
The patterns do not take source names or outgoing \modify{} edges into account; for example, \word{cat} in Fig.~\ref{fig:copulagraphs} would still exhibit the \pseven\ pattern if it had an incoming \app{\src{o}} edge instead of \app{\src{s}}, or if it were modified.} 
The \pzero~pattern corresponds to elements not present in the trees (Fig.~\ref{fig:patterns:zero}). If the word has an incoming \modify{} edge and no outgoing \app{} edges in its \lntree, like determiners in DM and PAS (Fig.~\ref{fig:catdoghouseAmdepPSD}, \word{the}), we assign it the \psix~pattern, shown in Fig.~\ref{fig:patterns:sixfour}. We have several patterns for words that are heads in the AM dependency tree, words that are either the root of the AM dependency tree or have an incoming \app{} edge (Fig.~\ref{fig:patterns}(\subref{fig:patterns:five}-\subref{fig:patterns:seven})). If such a head has no arguments of its own, i.e.~no outgoing \app{} edges, we use the \pseven~pattern in Fig.~\ref{fig:patterns:seven}. This pattern applies most commonly to nouns (e.g.~\word{cat} as \G{cat} in Fig.~\ref{fig:am:consts} or the nouns in Figs.~\ref{fig:copulagraphs} and \ref{fig:catdoghouseAmdep}). If the head has one or two arguments (= outgoing \app{} edges), as seen in Fig.~\ref{fig:patterns}(\subref{fig:patterns:five},~\subref{fig:patterns:onethree}), we assign it the \pfive~or \pone~pattern respectively. Intransitive and transitive verbs typically have these respective patterns; we find that the coordination structures of PAS and PSD in Fig.~\ref{fig:catdoghouseAmdep} show the \pone~pattern as well.

If a modifier (here defined as a word with an incoming \modify{} edge in the \lntree s) itself has an argument (an outgoing \app{} edge), we assign it the \ptwo~pattern (Fig.~\ref{fig:patterns:two}). This pattern is typical for prepositions in DM and PAS. We also find the following distinctions useful: sometimes a modifier, such as the temporal auxiliary \word{are} in the PAS analysis in Fig.~\ref{fig:catdoghouseAmdepPAS}, carries additional sources with it that create reentrancies at the time of modification. We use the \pfour~pattern (Fig.~\ref{fig:patterns:sixfour}) instead of \psix~for such complex modifications. When a head with two children has an annotation that will lead to a reentrancy between the children, such as the copula \word{is} with the constant \G{be} in Fig.~\ref{fig:am:consts}, we use the \pthree~pattern rather than \pone. Besides copula in PSD, all MRs have this pattern for control verbs. We summarize all other patterns as \peight.

Examining our pattern signatures, we find that about half of the nodes have the same pattern in all three graphbanks: for example, adjectives often display the signature \patterntriple{\psix}{\psix}{\psix} and nouns \patterntriple{\pseven}{\pseven}{\pseven}. These aligned pattern signatures represent a priori shared compositional structure across graphbanks, typically in the form of open-class words and predicate-argument structure. In contrast, Table~\ref{tab:patternSignatures} shows the most common pattern signatures with differences between the MRs. These pattern signatures map quite precisely to linguistic phenomena, specifically those involving closed-class functional words.



\begin{table}
    \centering
    \begin{tabular}{rclS[table-format=5.0]S[table-space-text-post = \%]}
        \toprule
        Rank & \shortstack[l]{Pattern signature\\(\patterntriple{DM}{PAS}{PSD})} & \shortstack[l]{Most commonly\\associated phenomena} & {Count} & {\shortstack[l]{Percentage\\of total}}  \\\hline
        1 & \patterntriple{\psix}{\psix}{\pzero} & Determiners & 49965 & 13.38\%  \\
        2 & \patterntriple{\pzero}{\psix}{\pzero} & Punctuation* & 45544 & 12.20\% \\
        3 & \patterntriple{\ptwo}{\ptwo}{\pzero} & Prepositions & 42395 & 11.35\% \\
        4 & \patterntriple{\pseven}{\pseven}{\psix} & NPs with prep.* & 30202 & 8.09\% \\
        5 & \patterntriple{\pzero}{\ptwo}{\pzero} & Prepositions & 22324 & 5.98\% \\
        8 & \patterntriple{\pzero}{\pfour}{\pzero} & Temporal auxiliaries & 14074 & 3.77\% \\
        10 & \patterntriple{\pfive}{\psix}{\psix} & Verbal Negation* & 7215 & 1.93\% \\
        14 & \patterntriple{\pzero}{\pone}{\pone} & Coordination & 5709 & 1.53\% \\
        20 & \patterntriple{\pzero}{\pthree}{\pone} & Copula (adjectival) & 2915 & 0.78\% \\
        21 & \patterntriple{\pzero}{\peight}{\peight} & Coordination & 2828 & 0.76\% \\
        \bottomrule
    \end{tabular}
    \caption{Most frequent pattern signatures we fix. Asterisks indicate that additional phenomena comprise a significant portion of the pattern. See Appendix \ref{app:patternsignatures} for complete list of pattern signatures we detect.
    }\label{tab:patternSignatures}
\end{table}


\subsection{The AM+ Algebra}\label{subsec:amplusalgebra}

Now that we can detect mismatches of the compositional structures
across graphbanks, we would like to modify these compositional
structures such that the AM dependency trees for all three graphs
contain the same edges. This is illustrated in
Fig.~\ref{fig:copulagraphs}(\subref{fig:copulaDMdecompNew}-\subref{fig:copulaPSDdecompNew}): All three trees have the same edges
and ignore the same tokens (namely, none); all three analyses agree on
the compositional structure, and all graphbank-specific differences
have been pushed into the graphs for the individual tokens.

However, there is one technical complication: These AM dependency
trees are invalid because some graphs have a root node which is also
marked with some other source (e.g.\ the one-node as-graph for ``is''
in Fig.~\ref{fig:copulaDMdecompNew}). This is necessary to make the copula
semantically vacuous -- we have a node for it in the AM dependency
tree, but there is no node in the DM as-graph to which the tree
evaluates -- but the AM algebra does not allow root nodes to also
have another source besides the root source.

We therefore extend the AM algebra to the \emph{AM+ algebra}, which is
exactly like the AM algebra but allows root nodes to contain additional
sources. In all other ways, the AM+ algebra is defined exactly like
the AM algebra, and the concept of an AM dependency tree generalizes
naturally to the AM+ algebra. Thus Fig.~\ref{fig:copulagraphs}(\subref{fig:copulaDMdecompNew}-\subref{fig:copulaPSDdecompNew}) are
valid AM dependency trees of the AM+ algebra.

The seemingly innocent step from the AM algebra to the AM+ algebra is
not without risks. On a technical level, the AM algebra is grounded in
the HR graph algebra \cite{CourcelleE12}. In this translation, root
nodes in the AM algebra are mapped to sources with a special name
(``root'') in the HR algebra, and a node in a graph of the HR algebra
may not be marked with more than one source at once. This is to
prevent situations where e.g.\ an ``apply'' operation could merge
together two nodes of the same graph. Thus we sacrifice the grounding
in the HR algebra and with it, certain technical guarantees.  From a
modeling perspective, the extra flexibility of the AM+ algebra means
that the same graph can be described with more terms of the AM+
algebra than with terms of the AM algebra. This can be useful -- in
fact, we rely on this flexibility here -- but it also means that the
choice of compositional structure must be made with care and, ideally,
in a linguistically informed manner. Making such a choice is the main
theme of the next section.

\section{Normalizing Compositional Structures Across Graphbanks}\label{subsec:unifyingkeyphenom}

We normalize the AM dependency trees using the AM+ algebra by applying transformations directly to the AM dependency trees of \lnt. The transformations described below are carefully designed not to change the graph the AM dependency trees evaluate to. We do this for all AM dependency trees by iterating over the sentences in the graphbanks and applying a transformation whenever its corresponding pattern signature matches. To make sure that we only apply a transformation when it matches the linguistic phenomenon we intend to address, we impose further restrictions based on POS tags and lemmas. The transformations are always applied in the same order (see Table~\ref{tab:SimilaritiesStep} in Section~\ref{sec:evaluation} below).
In the remainder of this section, we describe in detail how the individual transformations work and how they relate to the pattern signatures and linguistic phenomena.

\subsection{Category 1: Ignored Elements}

A common difference among the SDP graphs is that PSD does not represent the meaning of some tokens -- the tokens are \sortof{ignored} in the sense that they do not have incident edges -- whereas PAS includes nearly all tokens in the graph; DM falls somewhere in the middle. Section~\ref{subsec:localpatterns} discussed one such example: in Fig.~\ref{fig:catdoghouseSDP} the determiner \word{the} is part of the PAS and DM graphs, but not PSD.
In the corresponding \lntree s, ignored tokens are not part of the dependency tree: they have no incoming edge (\word{the} in Fig.~\ref{fig:catdoghouseAmdep}). Our fixes for these elements thus relies on creating vacuous lexical as-graphs that do not change the graph when combined. We address the following cases, which account for roughly 30\% of the divergent phenomena detected.

\paragraph{Determiners: \patterntriple{\psix}{\psix}{\pzero}.} As seen in Section \ref{subsec:localpatterns}, determiners show up as modifiers in the AM dependency trees for DM and PAS, but not in PSD, yielding the \patterntriple{\psix}{\psix}{\pzero} signature. 
\newline\textbf{Our fix:} We can modify the AM dependency tree for PSD to match the edges of the other AM trees (Fig.~\ref{fig:copulaPSDdecompNew}): we use a semantically vacuous constant for the determiner \word{the} that consists of only an unlabeled node, marked with a \src{det}-source and as the root at the same time. Performing the \modify{\src{det}} operation with this constant does not change the graph of the noun (here \word{cat}), and thus the AM dependency tree in Fig.~\ref{fig:copulaPSDdecompNew} still evaluates to the graph in Fig.~\ref{fig:copulaPSDgraph}. This construction makes sense: the \modify{\src{det}} operation reflects a relation on the sentence surface, and the empty determiner constant for PSD reflects the fact that the information carried by the determiner is ignored in the deeper PSD graph structure.

Note that since this empty lexical as-graph does not change the graph, we could attach it anywhere and still obtain the same graph as evaluation result. In other words, just looking at the PSD graph is not enough to tell us where the determiner should attach. We attach the new vacuous PSD determiners to the token where the determiner attaches in DM.
\newline\textbf{Further restrictions:} To make sure we exactly fix determiners this way, the token belonging to the local pattern must have the POS tag \pos{DT}. This covers 95\% of the pattern.

\paragraph{Punctuation, Particles \& Co.: \patterntriple{\pzero}{\psix}{\pzero}.} PAS is the only SDP MR to broadly annotate tokens that are mostly syntactic markers. This pattern includes punctuation, infinitival \word{to}, passive \word{by} and particles of particle verbs.
\newline\textbf{Our fix:} As for determiners, we add empty modifiers, here to DM and PSD.
\newline\textbf{Further restrictions}: POS tag indicates punctuation or is \pos{IN}, \pos{TO} or \pos{RP} (covers 97\% of pattern).

\paragraph{Temporal and Aspectual Auxiliary Verbs: \patterntriple{\pzero}{\pfour}{\pzero}.} An example of this is \word{are} in Fig.~\ref{fig:catdoghouseAmdep}. These auxiliaries are annotated only in PAS, where the \lntree s (Fig.~\ref{fig:catdoghouseAmdepPAS}) use complex modification to create the reentrancy observed in Fig.~\ref{fig:catdoghouseSDP:PAS}.
\newline\textbf{Our fix:} While the \pfour~pattern is different from the \psix~pattern as far as the resulting graph is concerned, both patterns simply have an incoming \modify{} edge when only looking at the AM tree edges. Thus, we can apply the same fix here as for the punctuation case. 
\newline\textbf{Further restrictions}: POS tag starts with \pos{V} (covers 82\% of pattern).


\paragraph{Copula: \patterntriple{\pzero}{\pthree}{\pone}.}
While all three SDP MRs treat the copula verb as a transitive verb in case of a nominal predicate \code{check this} (e.g.\ \sentence{Cats are mammals}), the treatment of copula with an adjectival predicate (e.g.\ \sentence{The cat is not lazy}, see Fig.~\ref{fig:copulagraphs}(\subref{fig:copulaDMgraph}-\subref{fig:copulaPSDgraph})) differs: 
DM treats this construction equivalent to adnominal adjectives (\word{cute dogs} in Fig.~\ref{fig:catdoghouseSDP:DM}), 
therefore the copula does not contribute directly to the graph, corresponding to the \pzero~pattern.
PSD and PAS, however, still encode the copula verb as a transitive verb thus differing from the structure of adnominal adjectives. PAS includes an edge from the adjective to the subject, resulting in a reentrancy in the graph and the \pthree~pattern in the AM tree. PSD does not include such an edge and expresses the relation between the adjective and the subject only via the copula. This results in the \pone~pattern in the AM tree.

While the \pone~pattern used in PSD differs from the \pthree~pattern employed by PAS, both MRs use the same AM dependency tree: the verb is the root and has outgoing \app{\src{s}} and \app{\src{o}} edges to the subject and the predicate respectively (Fig.~\ref{fig:copulagraphs}(\subref{fig:copulaPASdecomp},~\subref{fig:copulaPSDdecomp})). Thus, only DM exhibits a different compositional structure.
\newline\textbf{Our fix:} We normalize the AM dependency trees by adding a vacuous lexical as-graph for the copula verb in DM. This lexical as-graph consists of one root node with \src{o}-source, requiring an \src{s}-source in its argument (see \word{is} in Fig.~\ref{fig:copulaDMdecompNew}). We remove the \app{\src{s}} edge between the adjective (\word{lazy}) and the subject (\word{cat}), and instead make the subject and adjective \app{\src{s}} and \app{\src{o}} children of the copula respectively. Further, we move the incoming dependency edge of the adjective to the copula, making it the new head of the phrase. When evaluating this new AM tree, first the adjective (\word{lazy}) is plugged into the \src{o}-source of the copula (\word{is}), which results in an as-graph identical to the lexical as-graph of the adjective. Then the \src{s}-source is filled, plugging the subject into the correct slot of the adjective.
Note that this solution allows us to use the same DM adjective graph fragment for adnominal and predicative uses.

In practice, the key to automating this transformation is to find the \app{\src{s}} edge between the adjective and the subject in DM, since this is the edge we need to replace with the above construction. To find it, we look at the \app{\src{s}} and \app{\src{o}} children of the PSD \lntree \ to identify adjective and subject, and use the \app{\src{s}} edge between them in the DM \lntree. However, such an edge does not always exist, for example in cases where subject or adjective are complex phrases rather than single tokens, and PSD and DM disagree on the heads of these phrases. In these cases we cannot apply the fix automatically; the fix works in 86\% of the cases.
\newline\textbf{Further restrictions:} \word{be} lemma in PSD; the two outgoing \app{} edges in PSD AM dependency tree must be \app{\src{s}} and \app{\src{o}}; and the \app{\src{o}} child of PSD must have a POS tag starting with \pos{JJ} (adjective). This covers 64\% of the pattern.

\subsection{Category 2: Headedness}

For some constructions, there is disagreement across MRs about which token is the head of the construction. One such construction is verbal negation (roughly 2\% of divergent phenomena detected), for which we present a pattern and transformation here:

\paragraph{Verbal Negation: \patterntriple{\pfive}{\psix}{\psix}.}  For negation of verbs, we find that PSD and PAS treat the negation as modifier, while DM usually considers the negation to be the head (see Fig.~\ref{fig:copulagraphs}(\subref{fig:copulaDMgraph}-\subref{fig:copulaPSDdecomp})).
\newline\textbf{Our fix:} We make the negation the head in the AM dependency trees for PAS and PSD graphs. 
Consider the example PSD dependency tree in Fig.~\ref{fig:copulaPSDdecomp} and its transformation in Fig.~\ref{fig:copulaPSDdecompNew} (the process is the same for PAS). In the transformation of this example, we first reverse the dependency edge between the negation \word{not} and negated word \word{is} and use $\app{}$ instead of $\modify{}$. We then shift the incoming dependency edge (here the \el{root} marker) of \word{is} to \word{not}, making \word{not} the head of the phrase.
To make this work, we also need to change the graph fragment for \word{not}. This is because in the actual PSD graph (Fig.~\ref{fig:copulaPSDgraph}), \word{is} is still the root. We can make \word{is} the root in the resulting graph, even though \word{not} is the root of the AM dependency tree, by shifting the root in the graph fragment for \word{not} from the lexical label to the \src{neg}-source. This ensures that after the \app{\src{neg}} operation fills \word{is} into the \src{neg}-source, the root of the resulting graph is actually at \word{is}.
This transformation resembles type raising in lambda calculus where functor and argument are exchanged.
\newline\textbf{Further restrictions:} Lemma is \word{\#Neg} or \word{never} in PSD. This covers 35\% of the pattern; other phenomena adhering to this pattern include some discourse connectors and adverbs.
\\


\indent Besides conceptual differences in headedness like negation, some phrases simply do not have an obvious head, e.g.~a date like \word{Nov. 29}. In such cases, the MRs also often disagree on whether \word{Nov.} or \word{29} is the head. We leave the automatic detection and transformation of such cases to future work.

\subsection{Category 3: Relational Elements}

As a third category, the SDP graphbanks differ in how they mark certain linguistic relations such as coordination or preposition. We observe two strategies: (i) the relation is marked using only an edge between the two more \sortof{contentful} elements, ignoring the functional word; or (ii) the relation is marked using a node for the functional word with outgoing edges to the more contentful elements. These decisions permeate through to the AM dependency trees. We address the two most common phenomena of this category here: prepositions and coordination, accounting for roughly 30\% of the divergent phenomena we detect. In both cases, we normalize the AM dependency trees towards strategy (ii).


\paragraph{Prepositions and Their Effects: \patterntriple{\ptwo}{\ptwo}{\pzero}, \patterntriple{\pzero}{\ptwo}{\pzero}, \patterntriple{\pseven}{\pseven}{\psix}.}
DM and PAS use strategy (ii) for most prepositions and encode them as nodes, whereas PSD uses strategy (i) and encodes them as edges. See the preposition \word{in} in Fig.~\ref{fig:catdoghouseSDP}, to which the PSD \el{LOC} edge corresponds. In the AM trees, these prepositions have the \patterntriple{\ptwo}{\ptwo}{\pzero} signature.  Note that in the PSD graph in this example, the \el{LOC} edge corresponds to this preposition instead, which we find to be a general rule: if a preposition is not a node in the graph, the relation it expresses is instead encoded as an edge. In the original decomposition of \lnt, the \el{LOC} edge here is attached to the lexical as-graph of \word{house} yielding the \psix~pattern, a common cause for the \patterntriple{\pseven}{\pseven}{\psix} signature. Thus, normalizing the preposition case addresses two of the most common pattern differences at once.
Some prepositions such as possessives (\word{of}) are encoded as edges in DM, too, yielding the \patterntriple{\pzero}{\ptwo}{\pzero} signature.
\newline\textbf{Our fix:} In the graphs that encode the preposition as an edge, we use that edge as the lexical as-graph for the preposition, such as the lexical as-graph for \word{in} in Fig.~\ref{fig:catdoghouseAmdepPSDnew}. This new lexical as-graph consists of only the \el{LOC} edge, with the root node at its origin, and sources \src{s} and \src{o}. At the same time, we remove that edge from the lexical as-graph that contained it before (here \word{house}). We then update the AM dependency edges accordingly, for example in the transformation from Fig.~\ref{fig:catdoghouseAmdepPSD} to Fig.~\ref{fig:catdoghouseAmdepPSDnew}, we remove the \modify{\src{mod}} edge from \word{playing} to \word{house} and instead connect the new constant for \word{in} with an outgoing \app{\src{S}} edge to \word{house} and an incoming \modify{\src{O}} edge from \word{playing}, creating the \ptwo~pattern now in PSD as well.

To find the edge in PSD in the general case, we use a similar strategy as we employed for copula: we look at the parent of the preposition in PAS (\word{playing} in the example) as well as the \app{}-child of the preposition (\word{house}), and consider the PSD AM tree edge between them (here \modify{\src{mod}}). We then identify the graph edge (here \el{LOC}) via the source of the AM tree operation: the \el{LOC} edge here is the edge from \word{house} to the \src{mod}-source. Such a matching PSD AM tree edge exists in 86\% of the cases; in the other cases PAS and PSD disagree on the heads of the preposition arguments and we cannot apply an automatic transformation.

Note that this transformation also changes the \psix~pattern at \word{house} to the more consistent \pseven~pattern. In total, we have changed a \patterntriple{\ptwo}{\ptwo}{\pzero} and a \patterntriple{\pseven}{\pseven}{\psix} signature to a \patterntriple{\ptwo}{\ptwo}{\ptwo} and a \patterntriple{\pseven}{\pseven}{\pseven} signature. In the \patterntriple{\pzero}{\ptwo}{\pzero} case, we employ the same strategy to change the AM trees for both DM and PSD.
\newline\textbf{Further restrictions:} \pos{IN} or \pos{TO} POS tag at the preposition (covers 99\% of \patterntriple{\ptwo}{\ptwo}{\pzero} and 66\% of \patterntriple{\pzero}{\ptwo}{\pzero}).


\paragraph{Binary Coordination: \patterntriple{\pzero}{\pone}{\pone}, \patterntriple{\pzero}{\peight}{\peight}.}


\begin{figure}
	\centering
	\begin{subfigure}[b]{.45\textwidth}
	\centering
	\begin{dependency}[sdpgraph]
		\begin{deptext}[column sep=1em]
			Mice \& run \& and \& hide \\
		\end{deptext}
		\deproot[edge unit distance=2ex]{2}{ROOT}
		\depedge[edge unit distance=1ex]{2}{1}{ARG1}  
		\depedge[edge unit distance=1ex]{4}{1}{ARG1}  
		\depedge[edge unit distance=.5ex]{2}{4}{\_and\_c} 
	\end{dependency}
	\caption{DM graph}\label{fig:coordCommonSubj:sdpdm}
	\end{subfigure}
	\begin{subfigure}[b]{.45\textwidth}
	\centering
	\begin{dependency}[sdpgraph]
		\begin{deptext}[column sep=3em]
			Mice \& run \& and \& hide \\
		\end{deptext}
		\deproot[edge unit distance=2ex]{3}{ROOT}
		\depedge[edge unit distance=1ex]{2}{1}{verb\_ARG1} 
		\depedge[edge unit distance=1ex]{4}{1}{verb\_ARG1} 
		\depedge[edge unit distance=1ex]{3}{2}{coord\_ARG1}  
		\depedge[edge unit distance=1ex]{3}{4}{coord\_ARG2}  
	\end{dependency}
	\caption{PAS graph}\label{fig:coordCommonSubj:sdppas}
	\end{subfigure}
	
	\begin{subfigure}[b]{.45\textwidth}
	\centering
	\begin{dependency}[amdep]
		\begin{deptext}[column sep=.5cm]
			Mice \& run \& and \& hide \\
			\& \& \& \\
		\end{deptext}
		\deproot[edge unit distance=2ex]{3}{root}
		\depedge[edge unit distance=2ex]{3}{1}{\app{\src{s}}}      
		\depedge[edge unit distance=2ex]{3}{2}{\app{\src{op1}}}    
		\depedge[edge unit distance=2ex]{3}{4}{\app{\src{op2}}}    
		\node (n1) [below of = \wordref{2}{1}] {\includegraphics[scale=.5]{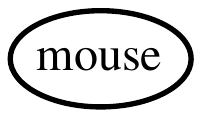}};
		\node (n2) [below of = \wordref{2}{2}] {\includegraphics{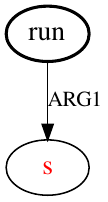}};
		\node (n3) [below of = \wordref{2}{3}] {\includegraphics{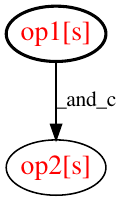}};
		\node (n4) [below of = \wordref{2}{4}] {\includegraphics{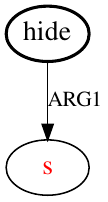}};
	\end{dependency}
	\caption{Normalized AM dependency tree for DM}\label{fig:coordCommonSubj:amdepdm}
	\end{subfigure}
	\begin{subfigure}[b]{.45\textwidth}
	\centering
	\begin{dependency}[amdep]
		\begin{deptext}[column sep=.5cm]
			Mice \& run \&[+6mm] and \&[+6mm] hide \\
			\& \& \& \\
		\end{deptext}
		\deproot[edge unit distance=2ex]{3}{root}
		\depedge[edge unit distance=2ex]{3}{1}{\app{\src{s}}}      
		\depedge[edge unit distance=2ex]{3}{2}{\app{\src{op1}}}    
		\depedge[edge unit distance=2ex]{3}{4}{\app{\src{op2}}}    
		\node (n1) [below of = \wordref{2}{1}] {\includegraphics[scale=.5]{./pics/pdf/mouse.pdf}};
		\node (n2) [below of = \wordref{2}{2}] {\includegraphics{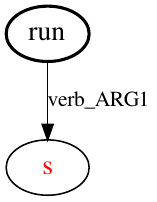}};
		\node (n3) [below of = \wordref{2}{3}] {\includegraphics[scale=1.5]{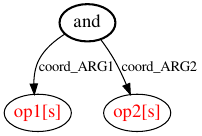}};
		\node (n4) [below of = \wordref{2}{4}] {\includegraphics{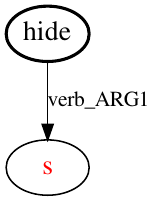}};
	\end{dependency}
	\caption{\lntree\  for PAS}\label{fig:coordCommonSubj:amdeppas}
	\end{subfigure}
	
	\begin{subfigure}[b]{.45\textwidth}
	\centering
	\includegraphics[scale=3]{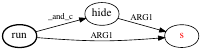}
	\caption{Partial result of (\subref{fig:coordCommonSubj:amdepdm}) after \app{\src{op1}} and \app{\src{op2}} operations}\label{fig:coordCommonSubj:partialdm}
	\end{subfigure}
	\begin{subfigure}[b]{.45\textwidth}
	\centering
	\includegraphics[scale=3]{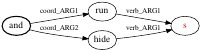}
	\caption{Partial result of (\subref{fig:coordCommonSubj:amdeppas}) after \app{\src{op1}} and \app{\src{op2}} operations}\label{fig:coordCommonSubj:partialpas}
	\end{subfigure}
	\caption{Coordination of two verbs with a common subject.
	}\label{fig:coordCommonSubj}
\end{figure}

To connect two conjuncts in coordination, such as \word{cats} and \word{dogs} in \sentence{cats and cute dogs} in Fig.~\ref{fig:catdoghouseSDP}, DM uses an edge (here \el{\_and\_c}; strategy (i)), while PAS and PSD make the conjunction \word{and} the head of the phrase with outgoing edges to the conjuncts (strategy (ii)).
Thus, both  PSD (as preprocessed by \lnt) and PAS show the same \pone~pattern in the \lntree s (Fig.~\ref{fig:catdoghouseAmdep}(\subref{fig:catdoghouseAmdepPSD},~\subref{fig:catdoghouseAmdepPAS})), while DM exhibits the \pzero~pattern.
If both conjuncts have a common argument, such as in Fig.~\ref{fig:coordCommonSubj:sdppas}, the source for the common argument is first passed up through the coordination before being filled with the argument, resulting in an additional \app{} child for the coordination (see Fig.~\ref{fig:coordCommonSubj}(\subref{fig:coordCommonSubj:amdeppas},~\subref{fig:coordCommonSubj:partialpas})). Thus, in such cases PAS and PSD \lntree s exhibit the \peight~pattern. In conclusion, binary coordination can have the  \patterntriple{\pzero}{\pone}{\pone} or \patterntriple{\pzero}{\peight}{\peight} signatures.

Binary coordination makes up approximately \code{check number} 86\% of all coordinations; we leave the more challenging case of three or more conjuncts to future work.
\newline\textbf{Our fix:} Similar to the preposition fix, we add a lexical as-graph for the conjunction token (e.g.~\word{and}) in DM containing the dedicated conjunction edge that was previously part of one of the conjuncts lexical as-graph (compare Fig.~\ref{fig:catdoghouseAmdepDM} and \ref{fig:catdoghouseAmdepDMnew}). The incident nodes use \src{op1}- and \src{op2}-sources to indicate the open positions for the conjuncts, and we connect the conjuncts as \app{\src{op1}} and \app{\src{op2}} children accordingly. One of the conjuncts was previously the head of the coordination in DM (here \word{cats}), we shift its incoming edge (here \app{\src{S}}) to the conjunction instead. The root of the conjunction's lexical as-graph is placed at the \src{op1}-source where \word{cats} will be placed, such that all outside graph edges attach correctly. If the conjuncts share a common argument like the subject \word{mice} in Fig.~\ref{fig:coordCommonSubj:sdpdm}, we add respective source annotations to the \src{op1}- and \src{op2}-sources (Fig.~\ref{fig:coordCommonSubj:amdepdm}). Fig.~\ref{fig:coordCommonSubj:partialdm} shows the partial result after evaluating the \app{\src{op1}} and \app{\src{op2}} operations in this example.
\newline\textbf{Further restrictions:} \pos{CC} POS tag, exactly two outgoing \el{*.member} edges in PSD (these are the coordination edges, exactly two ensures binary coordination). This covers 89\% of the two patterns.

\section{Evaluation}\label{sec:evaluation}
We evaluate the quality of the normalized AM dependency trees by measuring cross-formalism similarity as well as their suitability for parsing on the SDP corpora of the 2015 shared task \cite{oepen-etal-2015-semeval}. We apply \lnt's preprocessing of coordination for PSD (see also the discussion in Section \ref{sec:discussion}). Due to the similarities of the graphbanks, it is very appealing to use multi-task learning (MTL), which shares a subset of the model weights across graphbanks in order to let the training data for one graphbank influence the prediction on another. We expect this to help particularly in scenarios where there is limited data for one graphbank but large amounts for others. 

\subsection{Structural Similarity}
Tables \ref{tab:similarites} and \ref{tab:SimilaritiesStep} summarize the structural comparison of the graphs and AM dependency trees, respectively. Note that the F-scores on the graphs don't take the edge labels into account. We additionally calculate the percentage of lexical as-graphs that change by graph formalism as a result of our normalization: 13.53\% for DM, .35\% for PAS, and 33.22\% for PSD.  

There is a consistent increase in similarity by at least 12.8 points F-score of the normalized AM dependency trees in comparison to the \lntree s over all metrics. The labeled F-score that disregards the source names is the upper bound on how much similarity can be achieved by systematically making the source names more uniform as well. 

Comparing the similarities of the normalized AM dependency trees to the similarities of the graphs, there is an even larger gain in the unlabeled, directed F-score of up to 52.7 points. Importantly, we only compute F-scores on the edges of the AM dependency trees, not on the as-graph constants.  The higher similarity in the tree structure is achieved primarily by moving discrepancies to the lexicon. In particular, this is true for edge directions and edge labels, which are specified in the lexicon, not in the compositional structure.

\begin{table}[]
    \centering
\scalebox{0.7}{

\begin{tabular}{ll|cc|cccccccccccccc|cc}
\toprule
& & \multicolumn{2}{c}{\lnt} & \multicolumn{2}{c}{$\Delta$Det} & \multicolumn{2}{c}{$\Delta$Aux} & \multicolumn{2}{c}{$\Delta$Prep} & \multicolumn{2}{c}{$\Delta$Coord} & \multicolumn{2}{c}{$\Delta$Copula} & \multicolumn{2}{c}{$\Delta$Neg} & \multicolumn{2}{c}{$\Delta$ Punct} & \multicolumn{2}{c}{All changes} \\
& & PAS & PSD & PAS & PSD & PAS & PSD & PAS & PSD & PAS & PSD & PAS & PSD & PAS & PSD & PAS & PSD & PAS & PSD \\
\midrule
\multirow{3}{*}{DM} & UF & 63.5 & 55.7 & 0.0 & 6.8 & 1.3 & 0.8 & 3.3 & 9.9 & 2.8 & 3.2 & 0.6 & 0.6 & 0.5 & 0.5 & 4.3 & 1.3 & \textbf{76.3} & \textbf{78.8} \\
& A/M F & 57.8 & 46.7 & 0.0 & 7.2 & 1.3 & 1.0 & 3.4 & 11.3 & 2.8 & 3.2 & 0.6 & 0.7 & 0.5 & 0.5 & 4.5 & 1.8 & \textbf{70.9} & \textbf{72.4} \\
& LF & 29.7 & 26.2 & 0.0 & 8.1 & 1.6 & 1.4 & 3.6 & 1.3 & 2.6 & 2.9 & 0.6 & 0.7 & 0.5 & 0.4 & 5.3 & 3.9 & \textbf{43.9} &\textbf{44.9} \\
\multirow{3}{*}{PAS} & UF & & 57.0 & & 5.9 & & 1.3 & & 13.4 & & 0.0 & & 0.0 & & 0.0 & & 4.4 & & \textbf{82.0} \\
& A/M F & & 50.1 & & 6.1 & & 1.4 & & 13.7 & & 0.0 & & 0.0 & & 0.0 & & 4.6 & & \textbf{75.9} \\
& LF & & 36.8 & & 6.7 & & 1.5 & & 14.2 & & 1.1 & & 0.0 & & 0.0 & & 4.9 & & \textbf{65.2} \\
\bottomrule
\end{tabular}
}
    \caption{Directed F-scores between AM dependency trees at different stages of normalization. UF is unlabeled F-score, A/M F is labeled F-score disregarding source names and LF is the usual labeled F-score. 
    }
\label{tab:SimilaritiesStep}


\end{table}

\subsection{Suitability for Parsing}
An important question to ask is if the normalization of the AM dependency trees has an impact on how accurate parsing is. In particular, we expect the high similarity to be beneficial for projecting from a large corpus to a smaller one. Therefore, we conduct parsing experiments in the usual high-resource scenario and also in a simulated low-resource scenario, for which we sampled 3 subsets of the training data with 100 instances each and used those subsets throughout all low-resource experiments.

We use the parser of \lnt{} but, in contrast, we use only those training instances for which we have AM dependency trees in \emph{all} formalisms, resulting in 29,200 training instances. 
Moreover, we don't use embeddings for lemmas because of data sparsity in the low-resource condition. We follow their way of performing MTL, that is, we share an LSTM over all tasks in addition to using task-specific ones. Scores for edges and lexical as-graphs are produced from single-layer feedforward networks that take the concatenation of the output of the shared and task-specific LSTM as input. MTL experiments use the full training data of the non-target formalism and the specified amount for the target task; no data beyond the SDP corpora was used for the MTL experiments. For full details of hyperparameters, see appendix \ref{appendix:hyperparams}.

Table \ref{tab:parsingResults} summarizes the results. MTL in the low-resource scenario is hugely beneficial and works better with the normalized AM dependency trees. We think that the slight to moderate decrease in performance of the normalized trees in the full data condition comes from the noise the normalization creates because not all phenomena are detected perfectly and sometimes we restrict fixes to specific instances only.

\begin{table}[t]
    \centering

\scalebox{0.87}{
\begin{tabular}{llccc|cccccc|cc}
\toprule
& & \multicolumn{5}{c}{\lntree} & & \multicolumn{5}{c}{Normalized AM dependency trees} \\ \cline{3-7} \cline{9-13}
\multicolumn{2}{c}{Training data} & \multicolumn{3}{c}{100} & \multicolumn{2}{c}{full} & & \multicolumn{3}{c}{100} & \multicolumn{2}{c}{full} \\ 
& & Single & MTL & $\Delta$ & Single & MTL & & Single & MTL & $\Delta$ & Single & MTL \\
\multirow{2}{*}{DM} & id F & 65.6\ci{1.4} & 72.2\ci{0.9} & 6.6 & 92.8\ci{0.1} & 92.9\ci{0.1} & & 66.1\ci{1.2} & 72.9\ci{0.2} & 6.8 & 92.3\ci{0.2} & 92.2\ci{0.1}  \\
& ood F & 63.3\ci{1.3} & 69.1\ci{1.4} & 5.8 & 88.9\ci{0.1} & 88.9\ci{0.1} & & 63.0\ci{1.2} & 69.5\ci{0.5} & 6.5 & 88.0\ci{0.3} & 87.7\ci{0.3}  \\
\multirow{2}{*}{PAS} & id F & 72.6\ci{1.2} & 81.5\ci{0.2} & 8.9 & 94.7\ci{0.1} & 94.6\ci{0.1} & & 72.5\ci{1.4} & 84.1\ci{0.8} & 11.5 & 94.6\ci{0.1} & 94.5\ci{0.1}  \\
    & ood F & 69.2\ci{2.0} & 78.9\ci{0.7} & 9.7 & 92.6\ci{0.1} & 92.6\ci{0.1} & & 69.0\ci{1.6} & 80.7\ci{1.0} & 11.7 & 92.5\ci{0.1} & 92.6\ci{0.1}  \\
\multirow{2}{*}{PSD} & id F & 51.1\ci{0.7} & 58.9\ci{1.2} & 7.8 & 81.2\ci{0.1} & 80.9\ci{0.3} & & 52.9\ci{1.8} & 60.2\ci{0.9} & 7.4 & 80.5\ci{0.1} & 80.3\ci{0.1}  \\
& ood F & 50.3\ci{2.2} & 58.2\ci{1.8} & 7.9 & 80.3\ci{0.1} & 80.1\ci{0.2} & & 51.6\ci{1.2} & 59.1\ci{0.8} & 7.5 & 79.7\ci{0.1} & 79.5\ci{0.1} \\
\bottomrule
\end{tabular}
}
    \caption{Means and standard deviations of labeled F-scores on test sets for varying amounts of training data and use of MTL. Computed over 3 data samples with replacement and different random initializations.}
    \label{tab:parsingResults}
\end{table}


\section{Discussion}\label{sec:discussion}


The methodology we detail in Sections \ref{sec:updating} and \ref{subsec:unifyingkeyphenom} allows us to normalize roughly 60$\%$ of differences across compositional structures of the SDP graphbanks. Our work establishes that a small set of linguistically-grounded transformations are quite powerful towards creating uniform compositional structure. In future work, we plan to automate our methodology to broaden the range of phenomena we are able to normalize. 

The individual phenomena that create inconsistencies across graph representations and their corresponding \lnt~AM trees predictably adhere to a Zipfian distribution (Appendix \ref{app:patternsignatures}). Currently, we do not account for the interaction of these phenomena, such as negation combined with auxiliary verbs (\sentence{Mary will/might/should not sleep}) or adjectival copula with coordination (\sentence{the tree is tall and green}). We find that the order of implementing our fixes as shown in Table \ref{tab:SimilaritiesStep} minimally impacts F-scores and plan to explore this in more depth. 
Automating our methodology will enable us to better address these complex interactions and shed further light on how they impact compositional structure.

Although the AM+ algebra we present to handle discrepancies in compositional structure is more flexible than the \lnt\ AM algebra, there are a few phenomena that still pose challenges. In particular, this applies to the way coordination is originally handled in PSD. The coordinate structure in Fig.~\ref{fig:catdoghouseSDP:PSD} represents the PSD graph after the (reversible) preprocessing of \lnt. In a PSD graph without preprocessing, there would be two \el{ACT-arg} edges, pointing to \word{cats} and \word{dogs}. The AM+ algebra cannot handle this original structure unless we allow the lexical as-graph for \word{playing} to have two \el{ACT-arg} edges, an analysis that is questionable from a syntactic point of view. 


A final challenge we leave to future work involves better integrating our methodology with cross-framework MR parsing and MTL in general. Once we normalize representations across AM dependency trees, our target annotation has to account for all distinctions the different formalisms make. This makes parsing more difficult but we also believe this is where MTL can be of most use. The results obtained in the simulated resource scenario (Section \ref{sec:evaluation}) are promising first steps towards MR parsing in the low-resource scenario that is often faced when developing new MRs. We expect that, by automating our methodology to capture a broader range of phenomena, this will not only have further positive effects for the similarity of the compositional structures of MRs but will also make it more accessible to the community and help in the annotation process of future graphbanks.

\section{Conclusion} 

We have shown how annotations from different graphbanks (specifically, DM, PAS, and PSD) can be normalized at the level of their compositional structure. We achieve this by updating the AM algebra to the AM+ Algebra, which allows more flexibility in adapting compositional structures across MRs. Our new methodology includes a focused study of design differences between formalisms based on specific linguistic phenomena and more general pattern-based graph analysis. By working at the compositional level of AM dependency trees, we are able to quantify mismatches between different graphbanks, systematically reshape these mismatches to make them more uniform across graphbanks, and subsequently increase the match between the compositional structures for the three graphbanks. Such work contributes to a broader effort to increase parallel MR parsing accuracy.

As it is difficult to isolate and manually address all linguistic and structural discrepancies between graphbanks, we plan to extend the methodology we present in this paper to a system that is able to automatically detect and normalize mismatches. One possibility for this work includes detecting compositional mismatches during the algorithmic decomposition process itself \cite{groschwitz2018amr}. 
We also want continue exploring how normalization across graphbanks can bolster results with multi-task-learning, particularly in low-resource scenarios. 
As part of this work, we plan to extend our work normalizing compositional representations of MRs to formalisms with design architectures quite different from those discussed here. 




\bibliographystyle{coling}
\bibliography{AMheuristics}

\appendix 

\section{Additional Pattern Signatures and Frequencies}\label{app:patternsignatures}
See Table \ref{tab:patternSignaturesFull}.

\begin{table}[htb]
    \centering
    \begin{tabular}{rclS[table-format=5.0]S[table-space-text-post = \%]S[table-space-text-post = \%]c}
        Rank & \shortstack[l]{Pattern signature\\(\patterntriple{DM}{PAS}{PSD})} & \shortstack[l]{Most commonly\\associated phenomena} & {Count} & {Percentage} & {\shortstack[l]{Cumulative \\percentage}} & \shortstack[l]{We\\fix} \\\hline
        1 & \patterntriple{\psix}{\psix}{\pzero} & Determiners & 49965 & 13.38\% & 13.38\% & \checkmark\\
        2 & \patterntriple{\pzero}{\psix}{\pzero} & Punctuation & 45544 & 12.20\% & 25.58\% & \checkmark\\
        3 & \patterntriple{\ptwo}{\ptwo}{\pzero} & Prepositions & 42395 & 11.35\% & 36.93\% & \checkmark\\
        4 & \patterntriple{\pseven}{\pseven}{\psix} & NPs with prep.* & 30202 & 8.09\% & 45.02\% & \checkmark\\
        5 & \patterntriple{\pzero}{\ptwo}{\pzero} & Prepositions & 22324 & 5.98\% & 51.00\% & \checkmark\\
        6 & \patterntriple{\psix}{\pseven}{\pseven} & Named Entities* & 15835 & 4.24\% & 55.24\% & --\\
        7 & \patterntriple{\psix}{\pseven}{\psix} & Named Entities* & 14234 & 3.81\% & 59.05\% & --\\
        8 & \patterntriple{\pzero}{\pfour}{\pzero} & Temporal auxiliaries & 14074 & 3.77\% & 62.82\% & \checkmark\\
        9 & \patterntriple{\psix}{\psix}{\pseven} & Named Entities* & 7844 & 2.10\% & 64.93\% & --\\
        10 & \patterntriple{\pfive}{\psix}{\psix} & Verbal Negation* & 7215 & 1.93\% & 66.86\% & \checkmark\\
        11 & \patterntriple{\pseven}{\pseven}{\pfive} & Named Entities* & 6670 & 1.79\% & 68.64\% & -- \\
        12 & \patterntriple{\pfive}{\pseven}{\pseven} & Named Entities* & 6595 & 1.77\% & 70.41\% & -- \\
        13 & \patterntriple{\pseven}{\psix}{\psix} & Named Entities* & 5714 & 1.53\% & 71.94\% & -- \\
        14 & \patterntriple{\pzero}{\pone}{\pone} & Coordination & 5709 & 1.53\% & 73.47\% & \checkmark\\
        15 & \patterntriple{\pseven}{\psix}{\pseven} & \$ signs & 5008 & 1.34\% & 74.81\% & -- \\
        16 & \patterntriple{\pzero}{\psix}{\pseven} & Relative pronouns & 3703 & 0.99\% & 75.80\% & --\\
        17 & \patterntriple{\pone}{\ptwo}{\pzero} & Subordinating conjs. & 3609 & 0.97\% & 76.77\% & -- \\
        18 & \patterntriple{\pfive}{\pfour}{\pzero} & Modal auxiliaries & 3431 & 0.92\% & 77.69\% & --\\
        19 & \patterntriple{\pzero}{\ptwo}{\pone} & Punctuation & 2952 & 0.79\% & 78.48\% & -- \\
        20 & \patterntriple{\pzero}{\pthree}{\pone} & Copula (adjectival) & 2915 & 0.78\% & 79.26\% & \checkmark\\
        21 & \patterntriple{\pzero}{\pone}{\peight} & Coordination & 2848 & 0.76\% & 80.02\% & \checkmark\\
        22 & \patterntriple{\pzero}{\peight}{\peight} & Coordination & 2828 & 0.76\% & 80.78\% & \checkmark\\
        23 & \patterntriple{\pfive}{\pseven}{\pfive} & NCP & 2712 & 0.73\% & 81.51\% & -- \\
        24 & \patterntriple{\pone}{\pone}{\peight} & Particle verbs & 2612 & 0.70\% & 82.21\% & -- \\
        25 & \patterntriple{\ptwo}{\psix}{\pzero} & Determiners* & 2234 & 0.60\% & 82.80\% & -- \\
        26 & \patterntriple{\pzero}{\psix}{\psix} & Adverbs* & 2119 & 0.57\% & 83.37\% & -- \\
        27 & \patterntriple{\pseven}{\pseven}{\ptwo} & Named Entities* & 1992 & 0.53\% & 83.90\% & -- \\
        28 & \patterntriple{\ptwo}{\ptwo}{\peight} & NCP & 1851 & 0.50\% & 84.40\% & -- \\
        29 & \patterntriple{\psix}{\ptwo}{\pzero} & Subordinating conjs. & 1845 & 0.49\% & 84.89\% & -- \\
        30 & \patterntriple{\pzero}{\pone}{\pzero} & Punctuation & 1842 & 0.49\% & 85.39\% & -- \\
    \end{tabular}
    \caption{Most frequent pattern signature differences across the three graphbanks. Asterisks indicate that additional phenomena comprise a significant portion of the pattern. NCP denotes `no common pattern' or patterns that lack a single most common corresponding phenomenon.
    }\label{tab:patternSignaturesFull}
\end{table}



\section{Hyperparameters}
\label{appendix:hyperparams}
We use the implementation of \lnt : \href{https://github.com/coli-saar/am-parser}{https://github.com/coli-saar/am-parser}.
The hyperparameters common to all experiment are collected in Table \ref{tab:hyper}. We train the parser for 100 epochs and pick the model with the highest performance on the development set after epoch 25. We perform early stopping with patience of 10 epochs.

We use the "large-uncased" BERT model as available through AllenNLP and don't fine-tune.

\begin{table}
	\centering
	\begin{tabular}{ll}
		\toprule
		Activation function in all MLPs & tanh \\
		Optimizer & Adam \\
		Learning rate & 0.001 \\
		\textbf{Epochs} & \textbf{100} \\
		\midrule
		\textbf{Dim of lemma embeddings} & \textbf{0} \\
		Dim of POS embeddings & 32 \\
		Dim of NE embeddings & 16 \\
		\midrule
		Hidden layers in all MLPs & 1 \\
		\midrule
		Hidden units in LSTM (per direction) & 256 \\
		Hidden units in edge existence MLP & 256 \\
		Hidden units in edge label MLP & 256 \\
		Hidden units in supertagger MLP & 1024 \\
		Hidden units in lexical label tagger MLP & 1024 \\
		\midrule
		Layer dropout in LSTMs & 0.3 \\
		Recurrent dropout in LSTMs & 0.4 \\
		Input dropout & 0.3\\
		Dropout in edge existence MLP & 0.0 \\
		Dropout in edge label MLP & 0.0 \\
		Dropout in supertagger MLP & 0.4 \\
		Dropout in lexical label tagger MLP & 0.4 \\
		\bottomrule
	\end{tabular}
	\caption{Common hyperparameters used in all experiments. Deviations from L'19 are highlighted with boldface.}
	\label{tab:hyper}
\end{table}
\paragraph{Single Task} We use a batch size of 16 sentences.

\paragraph{MTL} In our MTL experiments, we have one LSTM per graphbank and one that is shared between the graphbanks. 
When we compute scores for a sentence, we run it through its graphbank-specific LSTM and the shared one. We concatenate the outputs and feed it to graphbank-specific MLPs.
We use a separate LSTM for the edge model (input to edge existence and edge label MLP) and the supertagging model.

In effect, we have two LSTMs that are shared over the graphbanks: one for the edge model and one for the supertagging model. All LSTMs have the hyperparameters detailed in Table \ref{tab:hyper}. We use a batch size of 16 for the target formalism and a batch size of 64 for all non-target formalisms. Each batch consists of instances from only one formalism.

\end{document}